\documentclass[sigconf]{acmart}
\acmSubmissionID{680}

\usepackage{booktabs} 

\usepackage{multirow} 
\usepackage{hyperref}
\usepackage[ruled]{algorithm2e} 

\SetAlFnt{\small}
\SetAlCapFnt{\small}
\SetAlCapNameFnt{\small}
\SetAlCapHSkip{0pt}

\newcommand{\yourevise}[1]{{#1}}
\def\papername{X-Portrait}

\usepackage{cuted}
\usepackage{capt-of}

\copyrightyear{2024}
\acmYear{2024}
\setcopyright{acmlicensed}
\acmConference[SIGGRAPH Conference Papers '24]{Special Interest Group on Computer Graphics and Interactive Techniques Conference Conference Papers '24}{July 27-August 1, 2024}{Denver, CO, USA}
\acmBooktitle{Special Interest Group on Computer Graphics and Interactive Techniques Conference Conference Papers '24 (SIGGRAPH Conference Papers '24), July 27-August 1, 2024, Denver, CO, USA}
\acmDOI{10.1145/3641519.3657459}
\acmISBN{979-8-4007-0525-0/24/07}

\begin{document}
\title{\papername: Expressive Portrait Animation with Hierarchical Motion Attention}

\author{You Xie}
\affiliation{%
 \institution{ByteDance}
 \country{USA}}
\email{you.xie@bytedance.com}
\author{Hongyi Xu}
\affiliation{%
 \institution{ByteDance}
 \country{USA}}
\email{hongyixu@bytedance.com}
\author{Guoxian Song}
\affiliation{%
 \institution{ByteDance}
 \country{USA}}
\email{guoxiansong@bytedance.com}
\author{Chao Wang}
\affiliation{%
 \institution{ByteDance}
 \country{USA}}
\email{chao.wang@bytedance.com}
\author{Yichun Shi}
\affiliation{%
 \institution{ByteDance}
 \country{USA}}
\email{yichun.shi@bytedance.com}
\author{Linjie Luo}
\affiliation{%
 \institution{ByteDance}
 \country{USA}}
\email{linjie.luo@bytedance.com}

\begin{strip}
	\centering
	\includegraphics[width=0.98\linewidth]{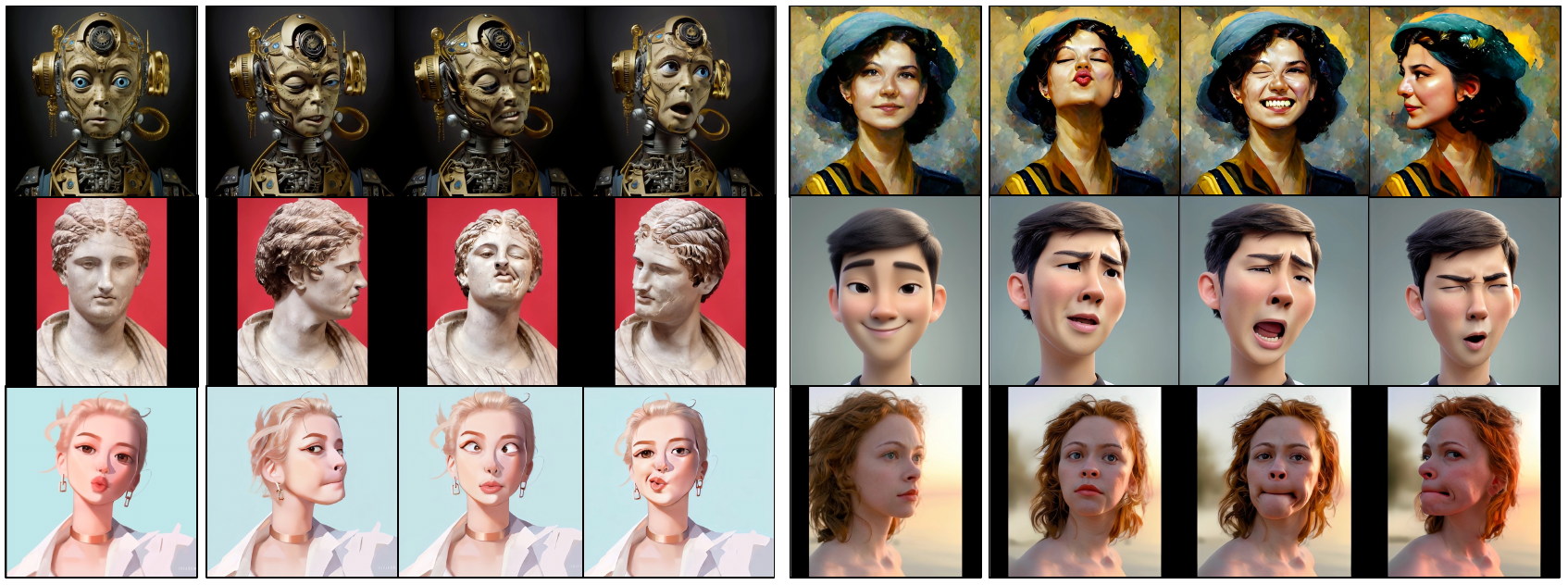}
	\centering
	\captionof{figure}{
 Given a single reference portrait (left column), \papername~ is capable of synthesizing compelling and expressive animations (right columns), covering large head pose changes and highly dynamic and detailed facial expression features from the input driving videos.
 Notably, \papername~ can faithfully preserve the identity information from the reference portrait while transferring the expression subtleties across a wide range of portrait styles. Please see the supplementary video for more dynamic results \yourevise{
 and \textbf{\href{https://github.com/bytedance/X-Portrait/tree/main}{\textcolor{blue}{here}} 
 for released code and models. \textcircled{c}George E. Koronaios, Dmitriy Ganin and Zacleonardi.
 }}  
}	\label{fig:teaser}
\end{strip}
\begin{abstract}
We propose \emph{\papername}, an innovative conditional diffusion model tailored for generating expressive and temporally coherent portrait animation. Specifically, given a single portrait as appearance reference, we aim to animate it with motion derived from a driving video, capturing both highly dynamic and subtle facial expressions along with wide-range head movements.  As its core, we leverage the generative prior of a pre-trained diffusion model as the rendering backbone, while achieve fine-grained head pose and expression control with novel controlling signals within the framework of ControlNet.  In contrast to conventional coarse explicit controls such as facial landmarks, our motion control module is learned to interpret the dynamics directly from the original driving RGB inputs. The motion accuracy is further enhanced with a patch-based local control module that effectively enhance the motion attention to small-scale nuances like eyeball positions. Notably, to mitigate the identity leakage from the driving signals,  we train our motion control modules with scaling-augmented cross-identity images, ensuring maximized disentanglement from the appearance reference modules. 
Experimental results demonstrate the universal effectiveness of \papername~across a diverse range of facial portraits and expressive driving sequences, and 
showcase its proficiency in generating captivating portrait animations with consistently maintained identity characteristics.
\end{abstract}

%
%

\ccsdesc[500]{Computing methodologies ~Animations}

%
%

\keywords{stable diffusion, ControlNet, generative model, portrait animation, talking head}

\maketitle
\section{Introduction}
Portrait animation, referring to the task of animating a single static portrait image using the head poses and facial expressions from a driving video likely featuring a different subject, has garnered growing attention owing to its versatile applications in video conferencing, visual effects and digital agents. In contrast to numerous prior works focused on talking head scenarios, we aim to transcend existing boundaries by synthesizing \emph{high-fidelity} head videos of \emph{in-the-wild} portraits in diverse styles, exhibiting \emph{highly dynamic} head poses and \emph{expressive} facial expressions. 

Commencing with the pioneering works~\cite{Siarohin_2019_NeurIPS,Siarohin_2019_CVPR}, portrait animation has predominantly been approached as a two-step generative process~\cite{zhao2022thinplate,Siarohin_2019_NeurIPS,wang2021facevid2vid,megaportraits}, involving image warping and rendering. Specifically, the identity features, encoded from the given source image, undergo an initial warping step with the motion flow calculated based on the disparities in expressions and head poses between the source and driving frame. Subsequently, these warped source features are fed into a generative decoder to synthesize the final animated frames with inpainted facial details and background. Despite substantial advancement in synthesizing high-fidelity talking heads, such methods exhibit limitations in both warping and rendering processes. The warping field, generated from landmarks~\cite{geng2018warp}, neural keypoints~\cite{wang2021facevid2vid} or latent code~\cite{wang2022latent}, often fails to capture subtle or extreme out-of-domain facial expressions. Additionally, due to the restricted capability of the decoder, the synthesized images are often limited in resolution and perceptual quality. Unnatural artifacts, such as blurriness and rigid shading variation, and facial distortions become apparent when animating out-of-domain portraits or driving with long-range head motions.



With the recent advent of text-to-image diffusion models~\cite{ho2020denoising,song2020denoising,song2020score,rombach2022high}, we have witnessed unprecedented diversity and stability in image synthesis demonstrated by large diffusion models pre-trained on billions of images, such as Imagen~\cite{saharia2022photorealistic} and Stable Diffusion (SD)~\cite{sd2022}. Consequently we aim to harness the generative capability of production-ready diffusion models, specifically SD1.5 in our work, for the task of portrait animation. Recently a line of work~\cite{hu2023animateanyone,chang2024magicpose,xu2023magicanimate,zhang2023dreamtalk} has approached this task as a problem of controlled image-to-video (I2V) diffusion. In particular, the appearance context from the source image is cross-queried by the self-attention layers of SD UNets, whereas the motion is controlled using explicitly derived semantic motion signals such as landmarks~\cite{chang2024magicpose}, skeleton~\cite{hu2023animateanyone,chang2024magicpose} and dense pose~\cite{xu2023magicanimate}, within the framework of ControlNet~\cite{zhang2023adding}. However, such explicit and coarse pose control signals do not fully convey the original expressions, and heavily rely on the robustness and accuracy of third-party pose detectors, hindering animation expressiveness and stability. Moreover, the controlling pose maps, derived from the driving frames, are not fully agnostic to the driving facial semantics (e.g., shape and ratio) and noticeably impact the source characteristics during cross-identity animation, referred to as appearance leakage here. 

In this work, we propose \papername, a novel zero-shot framework that leverages image diffusion priors for expressive portrait animation, showcasing superior perceptual quality, motion richness, identity preservation, and domain generalization (Fig.~\ref{fig:teaser}). Building upon prior controlled image-to-video works~\cite{hu2023animateanyone,chang2024magicpose,xu2023magicanimate}, we innovate the pose control scheme to mitigate the aforementioned issues, namely expressiveness loss and appearance leakage. To fully retain the driving head poses and facial expressions, we argue that the network should interpret the motion directly from the original driving images, without resorting to any intermediate motion representation. Subsequently, we design our pose ControlNet~\cite{zhang2023adding} using the RGB images as conditional inputs. However, standard motion transfer is typically trained on videos in a self-driven scheme  where the source and driving share the same identity. To avoid merely copying the appearance and structure directly from the conditional pose image by the model, we employ a refined state-of-the-art motion transfer network~\cite{wang2021facevid2vid} to generate \emph{cross-identity} training image pairs. Our cross driven training scheme simultaneously mitigates appearance leakage, enabling direct portrait animation during inference without any pre-processing. To further enhance the derivation of subtle facial expressions at nuanced scales, we deploy an auxiliary ControlNet~\cite{zhang2023adding} to guide the conditional motion attention to local facial movements especially around the eyes and mouth.

We extensively evaluate \papername~across our challenging benchmarks, comprising hundreds of driving videos that showcase a wide range of emotions, facial expressions, and head poses. \papername~demonstrates universal effectiveness for out-of-domain portraits and motions without any fine-tuning. Our method surpasses state-of-the-art portrait animation baselines both quantitatively and qualitatively, achieving superior visual fidelity, identity resemblance and motion accuracy. 
The contributions of our work can be summarized as follows:
\begin{itemize}
\item A novel zero-shot portrait animation method that extends pretrained Stable Diffusion with fine-grained control of head poses and facial expressions. 

\item A novel implicit motion control scheme trained with cross-identity scaling-augmented images, effectively retaining driving motion context to the fullest extent and simultaneously mitigating appearance leakage during inference. 

\item Enhanced interpretation of subtle facial expressions with guided motion attention to local facial movements. 

\item Demonstration of compelling fine-tuning-free portrait animation results with a diverse set of portraits and widely collected expressive driving sequences.
\end{itemize}

\begin{figure*}
	\centering
	\includegraphics[width=0.99\linewidth]{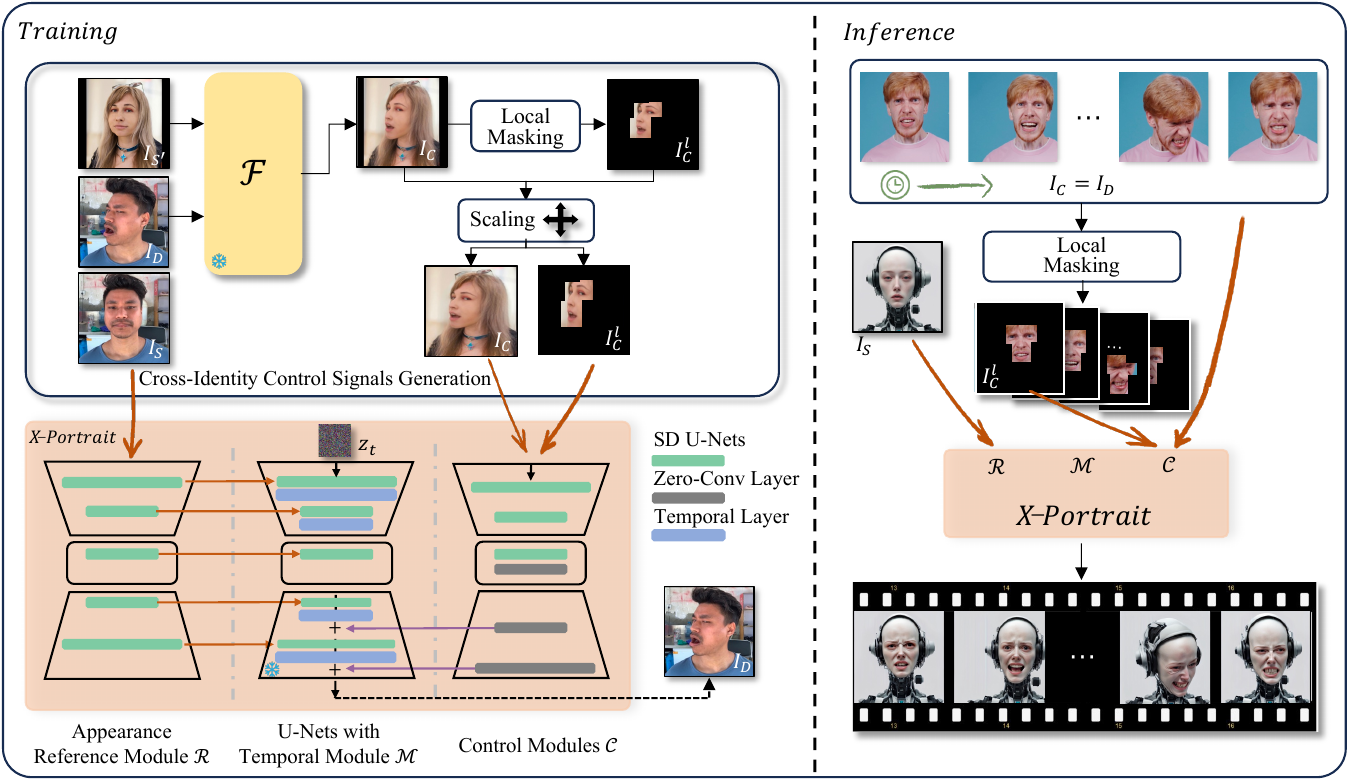}
	\centering
	\caption{\textbf{Overview of \papername~.} For the task of portrait animation, \papername~ leverages a frozen pre-trained LDM as a rendering backbone, and incorporates three auxiliary trainable modules for disentangled control of appearance $\mathcal{R}$, motion $\mathcal{C}$ and temporal smoothness $\mathcal{M}$. Specifically, $\mathcal{R}$ extracts the source appearance and background context from a reference image $I_S,$ and $\mathcal{C}$ derives the motion of head pose and facial expression from a driving frame $I_D.$ During training, we leverage a pre-trained network $\mathcal{F}$ to generate cross-identity control images $I_C$ as conditional input to our control modules $\mathcal{C}.$ To better capture subtle expressions, we enhance the attention to the local detailed facial movements with an additional masked control image $I_C^l.$ Both $I_C$ and $I_C^l$ are subject to random heterogeneous scaling for mitigation of appearance leakage from the drivings. For inference, we animate a source portrait directly with the video frames without any pre-processing, enabling expressive and robust animation with strictly maintained identity resemblance. \yourevise{\textbf{\textcircled{c}Tima Miroshnichenko.}}}  
    \label{fig:overview}
\end{figure*}
\section{Related Work}

\subsection{GAN-based Portrait Animation}
A majority of methods for portrait animation leverage Generative Adversarial Networks (GANs) to learn motion dynamics in self-supervised ways. The pioneering works\cite{Siarohin_2019_CVPR,Siarohin_2019_NeurIPS} consists of two stages, i.e., warping and rendering. They utilize sparse neural keypoints to estimate head motion, with which the encoded source identity features are warped and fed into a generative decoder to synthesize animated frames with inpainted facial details and background. Following this framework, a line of approaches have focused on enhancing the representations for warping estimation, such as 3D neural keypoints (Face Vid2Vid~\cite{wang2021facevid2vid}), depth(DaGAN~\cite{hong2022depth}) and thin-plate splines (TPS ~\cite{siarohin2021motion}). \yourevise{Additionally, ReenactArtFace \cite{10061279} employed a 3D morphable model to rig the expressions and poses, whereas ToonTalker \cite{gong2023toontalker} leveraged a transformer architecture to facilitate the warping process when dealing with cross-domain datasets.}
 To enhance rendered image quality, MegaPortraits~\cite{megaportraits} advances the generator to megapixel resolution by leveraging high-resolution image data. FADM~\cite{1020894} proposes a coarse-to-fine animation framework, employing a diffusion process to enrich facial details over the initial results generated by FOMM~\cite{Siarohin_2019_NeurIPS} and Face Vid2Vid~\cite{wang2021facevid2vid}. In addition to video reenactment, there have also been advancements embracing diverse driving signals, such as 3D facial prior~\cite{deng2020disentangled,Khakhulin2022ROME,sun2023next3d,DECA:Siggraph2021,10203414} and audios~\cite{guo2021adnerf,10.1145/3550469.3555393}. Nevertheless, these methods are reliant on explicit source features warping, with a primary focus on talking scenarios. This paper aims to generate high-fidelity expressive animations over portraits with diverse domain styles. 

\subsection{Diffusion-based Portrait Animation}
\yourevise{Recently Diffusion Probabilistic Models (DMs)~\cite{ho2020denoising,song2020denoising,song2020score} have achieved superior performance in various generative tasks,} such as image~\cite{saharia2022photorealistic}, video~\cite{blattmann2023stable}, and multi-view renderings~\cite{liu2023zero1to3, liu2023one2345,gu2023diffportrait3d}. 
Latent diffusion models~\cite{rombach2021highresolution} have further advanced this domain with reduced computational costs by operating the diffusion step in a lower-dimensional latent space. Relevant to our task of portrait animation, pre-trained diffusion models~\cite{saharia2022photorealistic,rombach2021highresolution} have been leveraged in many image to video (I2V) works. Notably, several studies~\cite{cao2023masactrl,lin2023consistent123,Zhangreference} highlight the effectiveness of fusing features of a reference image into the self-attention blocks in the LDM UNets, facilitating image editing and video generation with preserved appearance context.  Additionally, the nominal work ControlNet~\cite{zhang2023adding} extends the LDM framework to controllable image generation with additive structural conditions from signals such as landmarks, segmentations and dense poses. With those, several concurrent works~\cite{hu2023animateanyone,xu2023magicanimate,chang2024magicpose} achieve state-of-the-art full-body portrait animations by seamlessly integrating the appearance and motion control as well as temporal attentions~\cite{guo2023animatediff,guo2023sparsectrl} with pre-trained UNets. 
However, all of their motion control signals, i,e, skeleton with/without facial landmarks ~\cite{hu2023animateanyone,chang2024magicpose} and dense pose~\cite{xu2023magicanimate}, may not fully capture the original motion and depend heavily on the accuracy of third-party detectors, potentially hindering the animation expressiveness and stability. 
In contrast, we innovate the control signals to enable driving the portraits directly with a video, without resorting to any intermediate explicit motion representations. 
\section{Method}

Given as few as a single static portrait $I_{S}$, the objective of our approach is to generate a head animation sequence $\{I_{S\rightarrow D_i}\}$ with a length of $q$,  conditioned on a driving video ${I_{D_i}}$, where $i=0,\ldots,q$ denotes the frame index. The generated image $I_{S\rightarrow D_i}$ is intended to preserve the identity characteristics and background content depicted in $I_{S}$ while accurately following the head pose and facial expressions featured in the driving frame $I_{D_i}$. Conventional portrait animation algorithms, such as Face Vid2Vid~\cite{wang2021facevid2vid} (denoted as $\mathcal{F}$), are typically trained as a frame reconstruction task over a large corpus of videos, where $I_{S}$ and $I_{D}$ are both sourced from the same subject. During inference, the reference and driving images could feature distinct identities to achieve motion transfer.

In this work, \papername~ leverages the powerful generative prior of production-ready Latent Diffusion Models, with disentangled control of appearance and temporal motion~\cite{zhang2023adding} (Section~\ref{sec:pre}). On top of it, we propose a novel cross-identity training scheme with the aid of $\mathcal{F}$ that enables direct and accurate driving motion derivation from $I_{D}$ (Section~\ref{sec:crossid}). We further refine the control granularity by guiding the motion attention to local driving image patches with an auxiliary ControlNet~\cite{zhang2023adding} as elaborated in Section~\ref{sec:local}. Lastly, we provide details on several training and inference techniques for enhancing the retention of identity features from $I_{S}$ (Section~\ref{sec:id}). Figure.~\ref{fig:overview} provides an overview for the model architecture with the training and inference pipeline. 

\subsection{Preliminaries}
\label{sec:pre}
\paragraph{Latent Diffusion Model.}The Diffusion Models (DM)~\cite{ho2020denoising,song2020denoising,song2020score} are generative models designed to synthesize desired data
samples from Gaussian noise $z_T \sim N(0,1)$ through $T$ denoising steps. Latent diffusion models~\cite{rombach2022high} are a class of diffusion models that operates in the latent space facilitated by a pretrained auto-encoder. 
We note that, unlike standard  Text-to-Image (T2I) models conditioned on textual inputs, our application derives scene context and motion constraints from $I_{S}$ and $I_{D}$ respectively without text description. Consequently, text control is intentionally excluded from our formulations. The model is then trained to learn the reverse denoising process with the objective,
\begin{equation}
    L_{ldm} =\mathbb{E}_{z_0,t,\epsilon \sim \mathcal{N}(0,1)} \bigg[ \Big\lVert \epsilon-\epsilon_\theta \big(z_t,t\big) \Big\lVert_2^2 \bigg],
\end{equation} 
where $\epsilon_\theta$ is a trainable UNet with intervened layers of convolutions (ResBlock) and self-/cross-attentions (TransBlock). 
\\
\paragraph{Portrait Reenactment with Diffusion.}
A line of  works~\cite{hu2023animateanyone,chang2024magicpose,xu2023magicanimate} have recently delved into the utilization of pre-trained LDMs for full-body human video reenactment. While exhibiting slight algorithmic variations, these frameworks generally share a similar structure, consisting of plug-and-play modules in conjunction with a frozen SD UNet. Notably, these modules encompass: 1) 
an \emph{Appearance Reference Module} $\mathcal{R}$ that extracts the identity attributes and background context from $I_{S}$ to be cross-queried by the self-attention blocks of the backbone UNets; 2) a \emph{Control Module} $\mathcal{C}$, typically in a form of ControlNet~\cite{zhang2023adding}, that establishes structural mappings between control signals $I_C$ and the output with additive attentions; 3) a \emph{Temporal Module} $\mathcal{M}$~\cite{guo2023animatediff}, intervened within UNet blocks with temporal transformers, that ensures cross-frame correspondence for maintaining temporal coherence. We follow the same framework structure the but innovate on the controls, focusing on meticulous transfer of the source identity attributes as well as driving facial expressions and head poses.

\begin{figure}[t]
	\centering
	\includegraphics[width=0.99\linewidth]{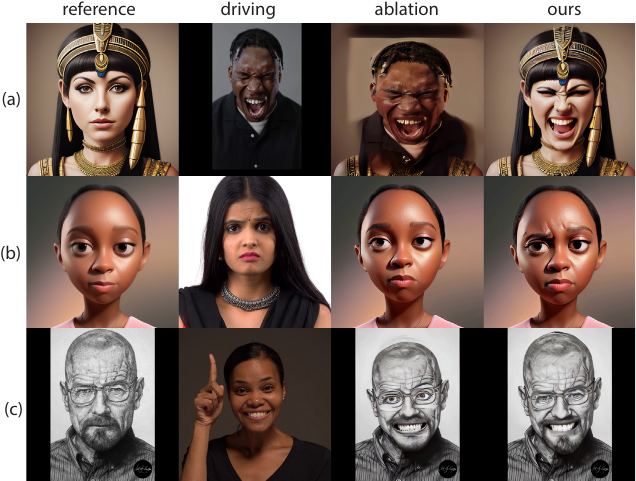}
	\caption{\textbf{Ablation.} (a) Training with scaling-augmented ground-truth driving image (with the same identity as reference) as motion condition results in severe identity appearance leakage from the driving. (b) Our local motion control module effectively enhances the capture of subtle and detailed local facial movements. (c) Random scaling with the conditional control images improves the identity preservation (note the head shape and eye sizes). \yourevise{\textbf{\textcircled{c}Archonom, Artbyhoussam, Cottonbro Studio and Ketut Subiyanto.}} }
	\label{fig:alba}
\end{figure}

\subsection{Implicit Cross-Identity Motion Control}
\label{sec:crossid}
To achieve control of facial expressions and head poses with image diffusion models, prior works typically employ a ControlNet~\cite{zhang2023adding} trained to condition image generation on facial landmarks, as showcased in e.g.,~\cite{chang2024magicpose}. The control module $\mathcal{C}$ is trained to reconstruct $I_D$ conditioned on the landmarks input extracted from the target $I_D,$ with $I_S$ as the input to the appearance reference module $\mathcal{R}.$
$I_S$ and $I_D$ are two random video frames during training, featuring the same subject. 
While effective at a coarse scale, such a control scheme induces several problems (Figure~\ref{fig:compare_all}), particularly when zoomed in on faces, as required in our task. 
Firstly,  the accuracy of the driving signals is heavily dependent on the precision of third-party detectors, such as OpenPose~\cite{openpose}. This dependence introduces jittered controls, motion ambiguity, and could result in corrupted animation when the detection fails, for example, due to face occlusion.  Secondly, the conveyance of strong emotions or subtle expressions often involves detailed facial movements, such as those in the teeth, eyeballs, eyebrows, and ajna. The animation expressiveness is significantly hindered by the coarse landmark representation, which may not capture the nuances demanded for accurate facial animation. Lastly, the driving landmarks are aligned with the face structure of targeted image $I_D,$ featuring the same subject as in $I_S$. Therefore under the self-driven training scheme, the network, as a short-cut, tends to copy the driving structure entangled with identity features such as facial shapes and ratios. As a result, undesirable identity drift to the driving subject occurs during cross-identity animation in inference. 

To address the aforementioned issues, we demand a novel conditional motion control that is entirely disentangled from the source identity features, while minimizing the loss of motion information at all scales, such as facial expressions and head poses. This motivates us to utilize the original driving RGB image $I_D$, featuring a different subject than $I_S$, as our conditional input. This aligns with our real-world applications as well, enabling the direct reenactment of the source image onto the driving video of a different identity. However, such image pairs with distinct identities but with aligned motions are not readily accessible for training.

Facilitated by a pre-trained portrait reenactment network $\mathcal{F}$ such as Face Vid2Vid~\cite{wang2021facevid2vid} in our work, we propose to synthesize cross-identity image pairs for the training of our control module $\mathcal{C}$. Specifically, two randomly selected video frames featuring the same subject are used, serving as the appearance reference $I_S$ and the reconstruction target $I_D$ respectively. However, instead of relying on facial landmarks from $I_D$, we employ $\mathcal{F}$ to generate an RGB control image $I_C = \mathcal{F}(I_{S'} \rightarrow I_D)$ as the conditional input to $\mathcal{C},$ where $I_{S'}$ is a frame randomly selected from a video with a distinct identity. In contrast to explicit extracted motion control signals, 
our cross-identity training scheme effectively instructs the control module $\mathcal{C}$ to implicitly derive the identity-disentangled motion from $I_C$. That being said, this mitigates appearance leakage from the driving signal, allowing direct application of the driving video for inference without third-party dependency. 
Notably, while degenerates with extreme expressions and long-range head motion, $\mathcal{F}$ offers reenacted control image $I_C$ of reasonable quality and motion accuracy for widely-distributed conversational scenarios. \yourevise{Even with limited perceptual quality, our control signal $I_C$ contains richer motion information than landmarks, which is sufficient for $\mathcal{C}$ to decipher the embedded motion structure effectively, enabling it to adapt and correlate to finer expressions and poses when provided with ground-truth motions for supervision.
Therefore, as illustrated in our experiments, our control module is able to establish implicit structural mapping between $I_C$ and $I_D,$ generalizing well to unseen expressions and head motions. Please find more discussion about control and generalization mechanism in Supplement Section B.}

\subsection{Enhanced Attention to Local Facial Movements}
\label{sec:local}
Our trained control module $\mathcal{C}$ offer a significant improvement over coarse landmarks in capturing head transformations and low-frequency facial expressions.
However, during inference,  we occasionally observe our control module fails to capture nuanced motion exhibited in the driving video, such as subtle nose wrinkles and intricate eyeball movements. These fine-grained motions, often overlooked by prior works, are crucial for conveying emotions and overcoming the uncanny valley artifacts in portrait animation.

Our control module $\mathcal{C}$, formulated as a ControlNet~\cite{zhang2023adding}, extracts structural features from the conditional image $I_C$ and integrates into the UNets via  skip connections during the denoising process. However, such additive conditional attention operates in the global image space, treating motion in every pixel with equal weight. This observation has motivated us to guide the control module to enhance localized attention specifically to critical facial regions, aimed at better animation realism and finer control granularity. 

We achieve motion control at nuanced scales by introducing an auxiliary ControlNet that conditions on a masked image $I^l_C$ revealing only patches around the eyes and mouth from $I_C$. \yourevise{Specifically, we detect landmarks of the control image for the eyes and mouth, using their centers to crop patches of $128\times128$  as local control signals $I^l_C$.}
While simple, this control branch effectively provides enhanced guidance to the UNet denoising, focusing solely on the local structure extracted from those cropped facial regions. Notably, before the training, we follow the same philosophy to fine tune the control image generator $\mathcal{F}$ with additional L1 and VGG~\cite{johnson2016perceptual} perceptual reconstruction losses on the local patches $I^l_C$. \yourevise{Our enhanced generator, named as \textit{Face Vid2Vid Plus}, helps capturing the subtle motions in the hierarchical conditional inputs $I_C$ and $I^l_C,$  benefiting the subsequent training of both control modules.}


\subsection{Identity Preservation}
\label{sec:id}
To ensure the preservation of source identity characteristics, we, to this end, have incorporated an appearance reference module $\mathcal{R}$ to derive appearance features from $I_S$ which are then concatenated into the UNet transformer blocks. Simultaneously, our cross-identity training scheme with reenacted conditional RGB inputs $I_C$ substantially mitigates the appearance leakage from the driving signals.  However, inherited from its self-supervised training, the pre-trained image reenactment generator $\mathcal{F}$ is not entirely free from appearance entanglement. Consequently, the facial attributes of $I_C$, especially in terms of face shape and the sizes of the eyes/mouth, are noticeably compromised by the target driving image $I_D$. That could result in slight identity drifts, especially when there are substantial differences in facial appearance between the source and driving. 

To alleviate the above mentioned issue, we adjust $I_C$ and $I^l_C$ with random heterogeneous scaling during training. The idea behind this modification is to induce slight face distortions and structure misalignments between the condition $I_C$/$I^l_C$ and the target $I_D$, forcing the network to rely on $I_S$ for identity features. We note that the scaling operations only impacts head shapes and do not modify the driving facial expressions and head poses.  \yourevise{While excessive induced misalignment can hinder the learning of the control modules, we empirically observe that 
random scaling factor within the range $[0.9,1.1]$
strikes a balance between identity preservation and motion expressiveness.} Additionally, during cross-driven inference, we further minimize the facial shape differences by applying an affine transformation (translation and scaling) over the entire driving sequence to align the head bounding box of the source and a selected driving frame. 

\begin{figure*}[t]
	\centering
	\includegraphics[width=0.99\linewidth]{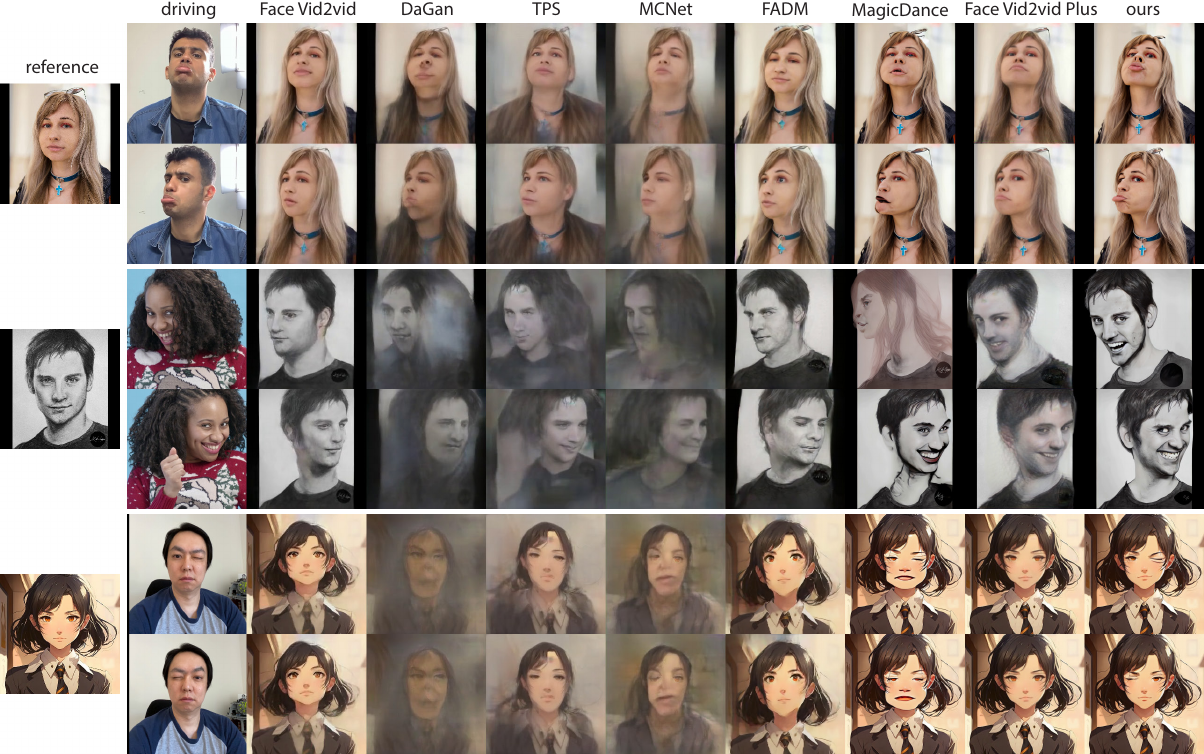}
	\caption{\textbf{Qualitative comparisons.}
 Among all the methods, \papername~ achieves the most accurate and robust transfer of both subtle and extreme facial expressions (e.g., pouting and single-eye blinks) and wide-range head translations and rotations, with precise identity resemblance (e.g., face shapes, eyes/mouth sizes) to the reference even with artistic styles (e.g., pencil drawing and anime). \yourevise{\textbf{\textcircled{c}Artbyhoussam, Midjourney.com and Katiin Bolovtsova.}}}
	\label{fig:compare_all}
\end{figure*}

With a single reference source image $I_S,$ only partial facial appearance is visible, and the network has to rely on the universal generative prior of LDM for inpainting unobserved facial regions when altering head poses or camera views. However, when more reference images are accessible, such as in a video, we can incorporate a more comprehensive appearance context without any network modification. Owing to our disentangled controls, by simply concatenating the multiple extracted appearance features into the UNets with $\mathcal{R}$, \yourevise{our framework can seamlessly fuse them and generate animations with better-retained identity attributes (more details in Supplement Section A).}

\section{Experiments}

\subsection{Implementation Details}
\paragraph{Dataset.}
\label{sec:dataset}
\yourevise{We train our model using an in-house dataset including monocular camera recordings of 42 expressions and 20-min talkings from 550 subjects in both indoor and outdoor scenes. All the data were processed with a cropped resolution of $512 \times 512$. Sequences of low quality were filtered out with~\cite{Hosu_2020}.}
All videos feature real subjects showcasing a diverse range of expressions and speeches in various scenes. 
For data processing, we follow the approach outlined in FOMM~\cite{Siarohin_2019_NeurIPS} but enlarge the cropping region to include the shoulders.
For evaluation, we collect 100 in-the-wild portraits~\cite{pexels,deviantart,midjourney} in various realistic or artistic depictions (2D/3D cartoon, anime, cyberpunk, oil painting, statue, wood, etc.), facial appearances (joker, elf, human-like robot, etc.), apparels (glasses, hat, robe, headphones etc.), and body poses (front and side). Additionally, we collect 200 licensed test videos with varying emotions, head poses, and facial expressions.

\yourevise{
\paragraph{Training and Inference.} 
We utilize SD 1.5 as our generative backbone and we freeze its weights during the entire training phases. \yourevise{Prior to training, the control and appearance reference modules are initialized using SD 1.5, whereas the motion module is initialized with the weights of AnimateDiff~\cite{guo2023animatediff}.} Our training was conducted in stages, where we sequentially plug in and train the modules $\mathcal{R}, \mathcal{C}$ and $\mathcal{M}$ . An AdamW optimizer is utilized with a learning rate of $10^{-5}$ to train all modules. 
Each module undergoes training with $30K$ steps with 16 video frames in each step.

 During inference, we leverage 
the prompt traveling strategy~\cite{tseng2022edge} to enhance temporal smoothness. With a frozen SD UNet, X-Portrait demonstrates inherent compatibility with the latent consistency model~\cite{luo2023latent}. This compatibility facilitates the efficient generation of a 24-frame animation within 30 seconds (10 steps) when executed on an A10 GPU. Notably, instead of denoising from  random Gaussian noise, 
we apply the forward diffusion process on source image $I_S$ into an initialized noise. Such generated noise adds a subtle level of structural guidance at the early denoising step, yielding improved consistency with reduced popping artifacts. We do not rely on $\mathcal{F}$ during inference.}

\begin{table*}[t!]
\centering
\caption{Quantitative comparisons of \papername~ with SOTA baselines in self and cross reenactment tasks, evaluated on an image resolution of $256\times 256.$}
\label{tab:quant_rec}
\scalebox{0.95}
{\begin{tabular}{lcccc|ccc}
\toprule
\multirow{2}{*}{Method}  & \multicolumn{4}{c}{\textbf{Self Reenactment}} & \multicolumn{3}{c}{\textbf{Cross Reenactment}} \\ \cmidrule(lr){2-5} \cmidrule(lr){6-8}
 & \textbf{L1}\ $\downarrow$  & \textbf{SSIM}\ $\uparrow$  & \textbf{LPIPS}\ $\downarrow$  & \textbf{FID}\ $\downarrow$ & \textbf{ID Similarity}\ $\uparrow$ &\textbf{Image Quality}\ $\uparrow$ & \textbf{Expression/Pose}\ $\downarrow$ \\
\midrule
Face Vid2vid~\citep{wang2021facevid2vid}   &0.042  &0.770&0.203&52.071 &0.615&36.222&0.100/4.72  \\
DaGAN~\citep{hong2022depth}  &0.054 &0.732 &0.234&64.297&0.280&35.537&0.111/5.10  \\
TPS~\citep{zhao2022thinplate}   &0.038 &0.801 &0.179&50.636 &0.508&34.904&0.079/3.94 \\
MCNet~\citep{hong23implicit}   &0.035 &0.811 &0.176&51.091&0.343&34.624& 0.094/3.82 \\
FADM~\cite{1020894}  &0.049 &0.727 &0.231&51.091 &0.635&34.640 &0.103/4.62\\
MagicDance~\citep{chang2024magicpose}   &0.045 &0.766 &0.172&30.383 &0.643&67.241 &0.099/7.04 \\
\midrule
Face Vid2vid Plus &0.034   &0.825  & 0.144&26.386&0.625&42.478&0.071/3.44 \\ 
\papername~  &\textbf{0.033} &\textbf{0.829} &\textbf{0.133}&\textbf{14.553}&\textbf{0.689} &\textbf{67.569} & \textbf{0.070/3.37} \\ 
\bottomrule
\end{tabular}}
\end{table*}

\subsection{Evaluations and Comparisons}
\begin{figure*}[t]
	\centering
	\includegraphics[width=0.99\linewidth]{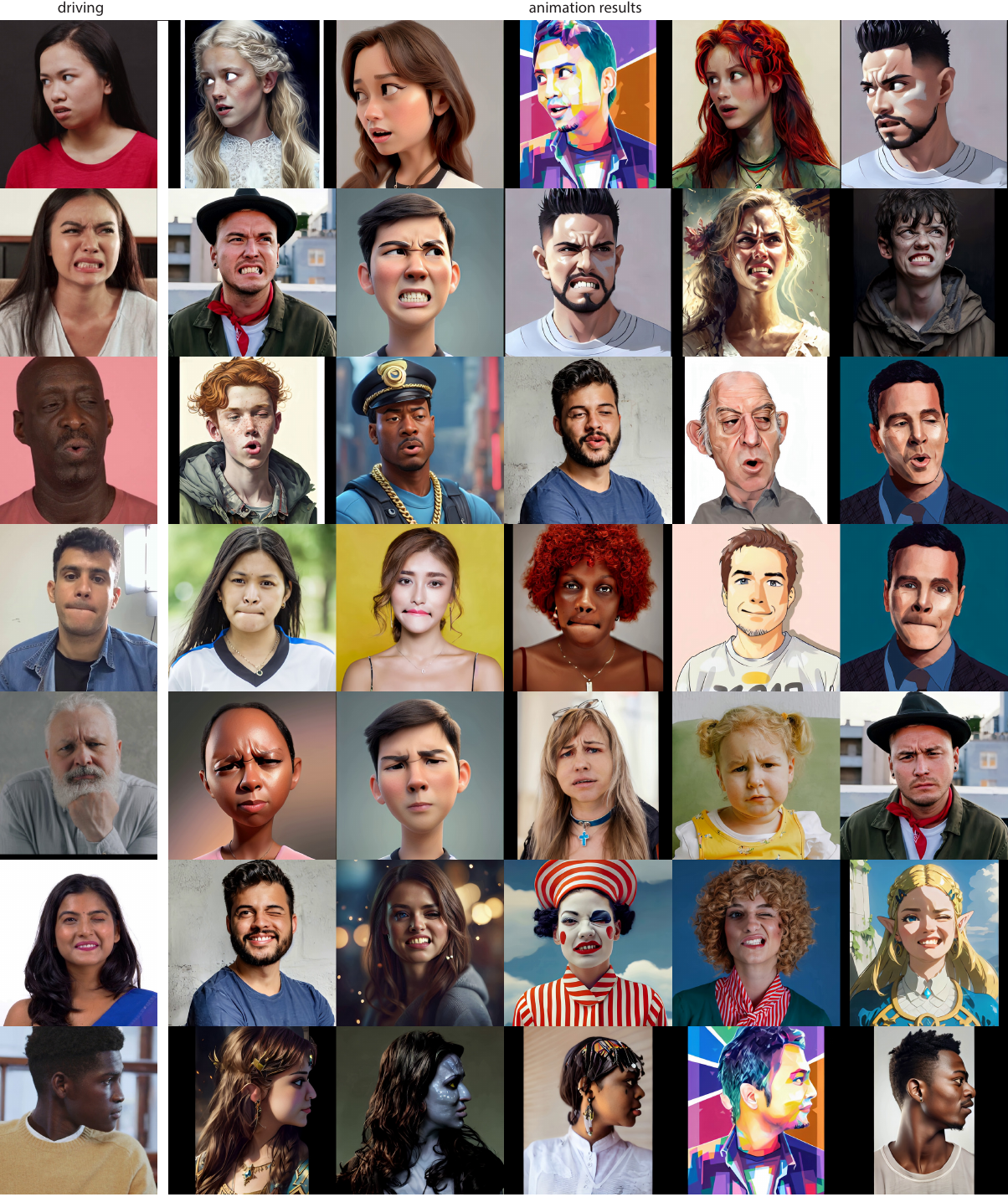}
	\caption{\textbf{More qualitative results.}The leftmost column shows the driving frame, and we show the animated results for multiple portraits by \papername. \yourevise{\textbf{\textcircled{c}Mart Production, Ketut Subiyanto, Alena Darmel, Gifing.com and Cottonbro Studio.} } }
	\label{fig:big}
\end{figure*}

\begin{figure*}[t]
	\centering
	\includegraphics[width=0.99\linewidth]{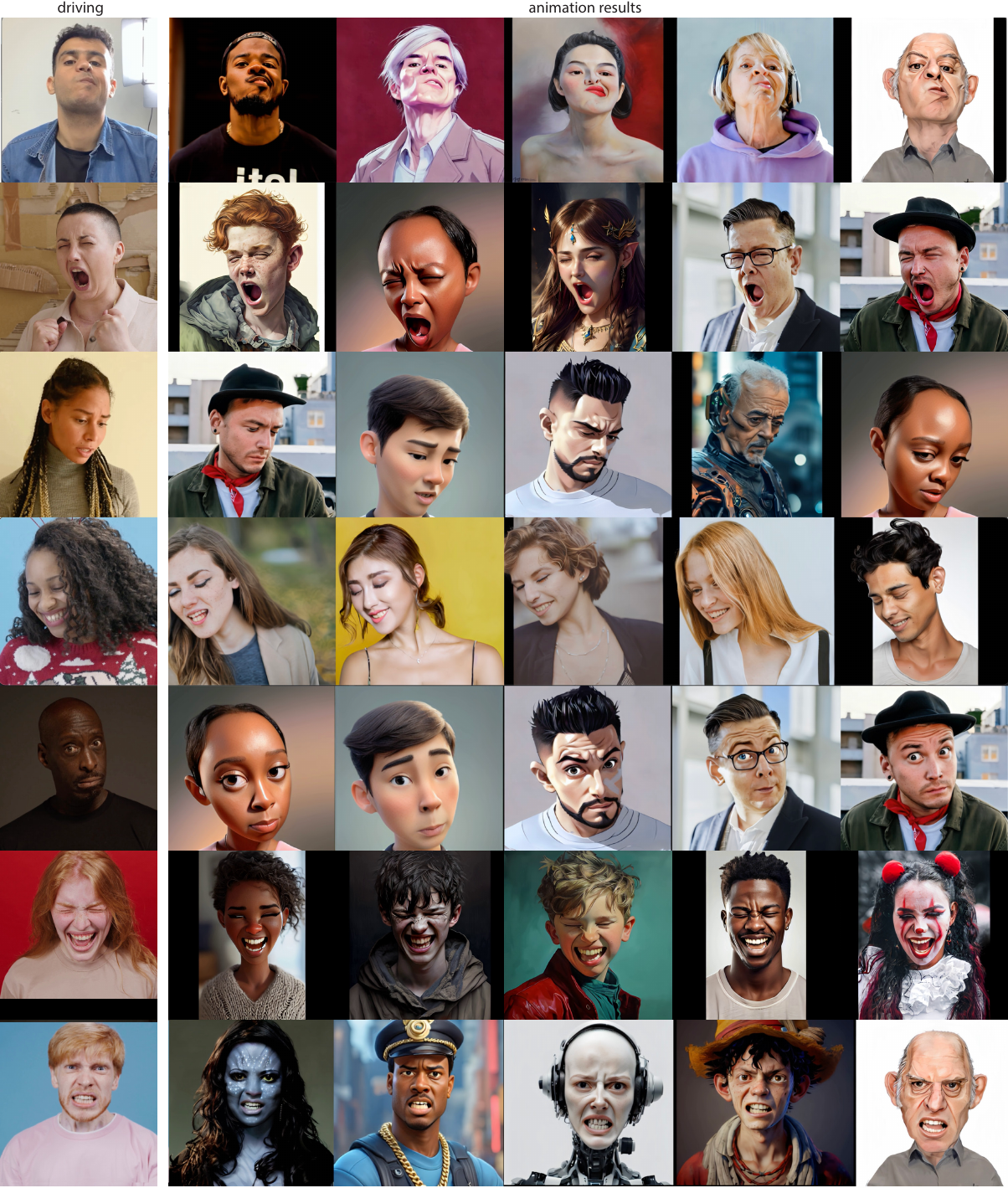}
	\caption{\textbf{More qualitative results.}The leftmost column shows the driving frame, and we show the animated results for multiple portraits by \papername. \yourevise{\textbf{\textcircled{c}Ketut Subiyanto, Tima Miroshnichenko, Karolina Grabowska, Alex Green, Katrin Bolovtsova.} } }
	\label{fig:big2}
\end{figure*}

Our method empowers the creation of captivating and highly expressive animations, demonstrating a diverse range of head motions (with rotations over 150 degrees) and facial expressions (frowning, crossed eyes, pouting etc) across both realistic, human-like, and style portraits (Figure~\ref{fig:teaser},~\ref{fig:big},~\ref{fig:big2}). \yourevise{Note that X-Portrait employs a reference module to effectively cross-query source appearance features, thereby establishing localized spatial correspondences between the input and output. Once trained, the model is able to generalize to out-of-domain appearances through its learned latent space, as exemplified by stylized portraits. }
Simultaneously, we maintain high identity resemblance to the given source image throughout the generated video. We compare our method with prior portrait animation works including state-of-the-art 
GAN based methods Face Vid2vid~\cite{wang2021facevid2vid} (and our enhanced version Face Vid2vid Plus), TPS~\cite{zhao2022thinplate}, DaGAN~\cite{hong2022depth}, MCNet~\cite{hong23implicit} and recent diffusion-based approaches like FADM~\cite{1020894} and MagicDance~\cite{chang2024magicpose}. We exclude MegaPortraits~\cite{megaportraits} from our comparisons as there is no public release.
For fair comparisons, we fine-tune all the baselines over the same dataset. Notably, for MagicDance~\cite{chang2024magicpose}, we use our own implementation and fine-tune the 68-landmark-based ControlNet from the public checkpoint. We assess their performances over both self and cross reenactments. All numbers are computed at the resolution of $256\times256$ due to the limited resolution  for most of the previous works.

\paragraph{Self Reenactment.} 
For each test video, we use the first frame as the reference image and generate the entire sequence where the subsequent frames serve as both driving image and the ground truth target. As shown in the numerical comparisons (Table~\ref{tab:quant_rec}), \papername~ consistently demonstrates superior image quality and motion accuracy over all the baselines. 



\paragraph{Cross Reenactment} As evidenced in our qualitative comparisons in Figure~\ref{fig:compare_all}, \papername~ demonstrates the best perceptual quality, motion precision and identity similarity. Despite being trained exclusively on real portrait animation, our method exhibits superior domain generalization capabilities to style portraits, whereas all GAN-based baselines suffer from severe blurriness for out-of-domain styles and under large head motions. For both extreme (e.g., pouting, top rows) and subtle expressions (e.g., single eye blinks, bottom rows), all the methods except ours fail to capture and transfer the nuances to the reference portraits. 
MagicDance~\cite{chang2024magicpose} yields unstable animations when landmarks detection fails (middle rows) and identity drifts when the source and driving have distinct facial features (bottom rows).


For quantitative assessment, given the absence of image ground truth, we employ three metrics to evaluate identity similarity, image quality, and expression and head pose accuracy, respectively. Specifically, the ArcFace score~\cite{deng2019arcface}, calculating the cosine similarity between source and generated images, is utilized to assess identity preservation. We employ a pre-trained network (HyperIQA~\cite{Su_2020_CVPR}) for image quality assessment. To evaluate the motion accuracy, we compare the L1 difference between the extracted facial blendshapes and head poses of the driving and generated frames using ARKit~\cite{arkit}. 
As reported in Table \ref{tab:quant_rec}, our method consistently outperforms all competitors by a good margin. 
\yourevise{Furthermore, a user study was carried out to affirm that X-Portrait significantly exceeds FaceVid2Vid Plus in terms of motion accuracy and expressiveness (more details in Supplement Section C).} 
Notably, by leveraging the SD prior, \papername~ and MagicDance surpasses the other methods by a substantial margin in image quality. However, compared to MagicDance, \papername~ shows superior identity preservation and motion accuracy, thanks to our novel hierarchical cross-identity scale-augmented controls.

\begin{table}[h]
\caption{Quantitative ablation.}
\begin{center}
\setlength{\tabcolsep}{1.5mm}
\renewcommand{\arraystretch}{1.1}
\begin{tabular}{l|cc}
\toprule
  Method  &ID~Similarity $\uparrow$  & Expression/Pose $\downarrow$ \\
\midrule
(a) w/o cross-id training & 0.015 & 0.037/1.58 \\
(b) w/o local control  &  0.731  & 0.077/3.69   \\
(c) w/o scaling &0.658& 0.070/3.26\\
\midrule
full model &0.689& 0.070/3.37 \\
\bottomrule
\end{tabular}
\end{center}
\label{tab:ablation} 
\end{table}

\subsection{Ablation Studies}
We ablate the efficacy of individual components by removing them from our full training pipeline, evaluated on cross reenactment synthesis.
Figure~\ref{fig:alba} (a) demonstrates the results when training the model naively with the driving frame as both the target and motion condition (self-driven training, even with our scaling strategy). In this scenario, the network tends to treat it as an image reconstruction task and merely copies both the identity and motion from the driving frames. Therefore, as shown in the quantitative evaluation(Table~\ref{tab:ablation} (a)),  while the expression accuracy is on par with our full pipeline, there is a significant decrease in identity resemblance.
Figure~\ref{fig:alba} (b) highlights the importance of our local control module in capturing subtle driving expressions. Excluding the local control module results in the absence of expression details, such as the asymmetric frowning,  aligning with the observation of decreased expression accuracy (Table~\ref{tab:ablation} (b)). We note that it achieves a slightly better identity similarity score due to the more rigid local facial movements. Furthermore, the source identity features are better maintained with our scaling augmented training strategy without which 
noticeable identity drift to the driving occurs, as
evidenced in Figure~\ref{fig:alba} (c) and Table~\ref{tab:ablation} (c). 


\yourevise{
\subsection{Limitations and Future Work}
\papername~ represents a general pipeline that improves the expressiveness of the base reenactment method $\mathcal{F}$. 
%
Nevertheless, due to representation limitation, $\mathcal{F}$ may not activate at all for cases with exceptionally challenging expressions. In such instances, these expressions are filtered out by $\mathcal{F}$ during training, posing challenges for~\papername~to learn the corresponding distribution, as illustrated in Figure~\ref{fig:limitations}.
\begin{figure}[t]
	\centering
	\includegraphics[width=0.99\linewidth]{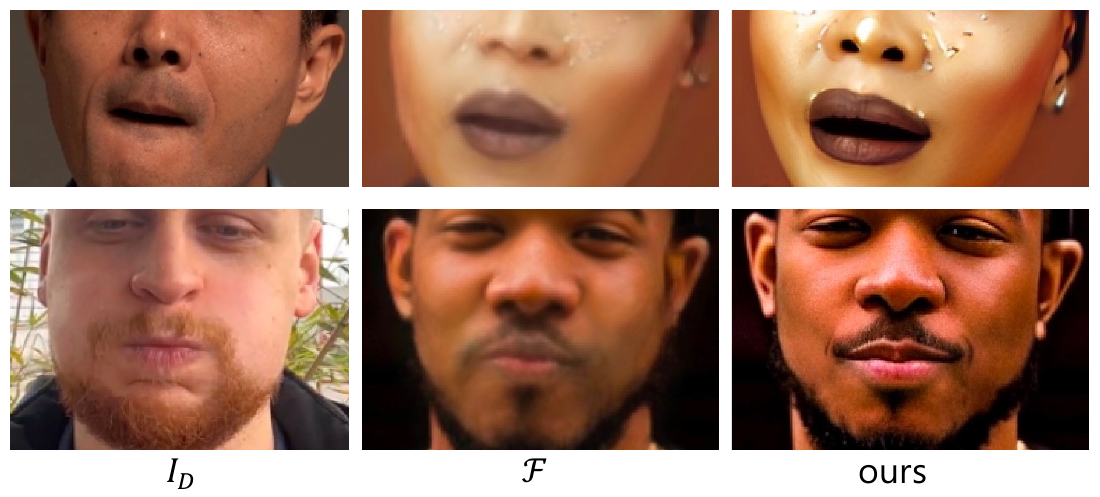}
	\caption{\yourevise{\papername~exhibits limited transfer of expressions when $\mathcal{F}$ completely fails to produce any correlated motion clues (e.g., turning lips inwards in the first row and puffing cheeks in the second row).}} 
	\label{fig:limitations}
\end{figure}
In the future, we aim to enhance the model's capabilities by animating gestures to further improve expressiveness. As a parallel objective, results quality could benefit from refined base diffusion models, e.g., the teeth region, and advanced spatiotemporal attentions could be explored to reduce noticeable jittering artifacts. 

}
\section{Conclusion}
We introduce \papername, a novel SD-based framework crafted for portrait animation, ensuring meticulous transfer of driving facial expressions and head poses. Our method excels with the incorporation of cross-identity driving inputs in training, facilitating a balanced achievement of motion expressiveness, identity preservation, and animation robustness. Additionally, we propose a local control module to accentuate the attention to detailed facial expressions that are subtle to capture but critical to emotion conveyance.  The showcased impressive performance of our model on generalized source portraits and driving motions validates its effectiveness.


\bibliographystyle{ACM-Reference-Format}
\bibliography{main}
\end{document}


\title{\papername: Expressive Portrait Animation with Hierarchical Motion Attention (Supplementary)}

\author{You Xie}
\affiliation{%
 \institution{ByteDance}
 \country{USA}}
\email{you.xie@bytedance.com}
\author{Hongyi Xu}
\affiliation{%
 \institution{ByteDance}
 \country{USA}}
\email{hongyixu@bytedance.com}
\author{Guoxian Song}
\affiliation{%
 \institution{ByteDance}
 \country{USA}}
\email{guoxiansong@bytedance.com}
\author{Chao Wang}
\affiliation{%
 \institution{ByteDance}
 \country{USA}}
\email{chao.wang@bytedance.com}
\author{Yichun Shi}
\affiliation{%
 \institution{ByteDance}
 \country{USA}}
\email{yichun.shi@bytedance.com}
\author{Linjie Luo}
\affiliation{%
 \institution{ByteDance}
 \country{USA}}
\email{linjie.luo@bytedance.com}

\maketitle
\yourevise{
\appendix
\section{Inference with multiple reference images}
\label{sec:xportrait_details}
\begin{figure}[t]
	\centering
	\includegraphics[width=0.95\linewidth]{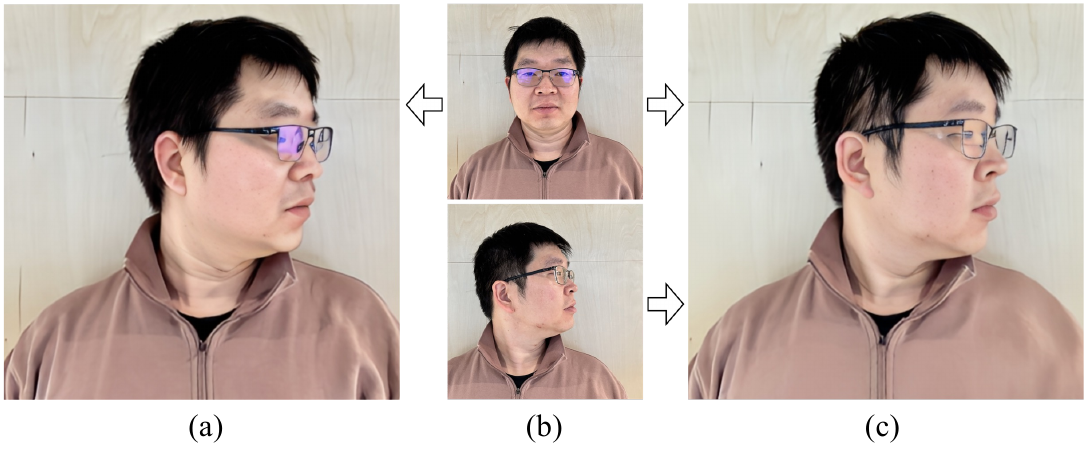}
	\caption{ \yourevise{(a) is the result of \papername~ with a single reference image from (b), while (c) with both reference images. \papername~ seamlessly accommodates multiple images as reference, producing animations with better captured personalized appearance traits in (c). Please find the differences in hair, ear and face shape.}}
	\label{fig:multisrc}
\end{figure}
\begin{figure}[t]
	\centering
	\includegraphics[width=0.95\linewidth]{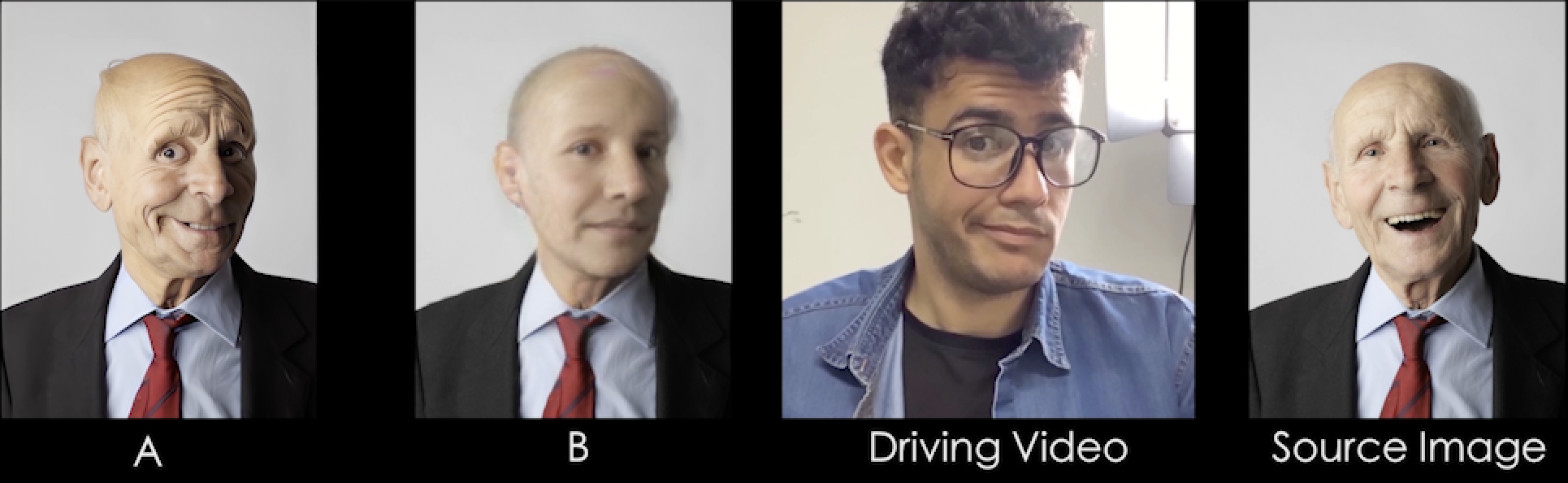}
	\caption{ \yourevise{User study example. A and B represent synthesized outputs from different methods.}}
	\label{fig:user_study}
\end{figure}

Features extracted by the appearance reference module $\mathcal{R}$ from the reference source image $I_S$ are cross-queried by the self-attention blocks within the backbone UNets. Consequently,~\papername~can be naturally extended to scenarios where multiple $I_S$ images are available. Utilizing more reference source images results in improved personalized appearance features, as illustrated in Figure~\ref{fig:multisrc}.


\section{Control and generalization mechanism}
\label{sec:control_mechanism}
\papername~employs a large pre-trained latent diffusion model (SD 1.5) as the foundational synthesis framework, leveraging its expansive latent space to generate portraits spanning diverse expressions and styles. Nevertheless, pinpointing the latent features corresponding to specific appearance and expression poses a significant challenge. X-Portrait utilizes a reference module to cross-query source appearance features, establishing localized spatial correspondences between the source and output. Once learned, the model is able to generalize to out-of-domain appearances from its latent space. Motion control employs a different mechanism with additive attentions injecting structural guidance towards U-Net denoising. The conditional control images, generated by FaceVid2Vid Plus, are not used as supervision targets. They are not required to be strictly photo-realistic and pixel-wise accurate as long as it provides some structural motion clues (embedded as high-dimensional spatial cross attentions).  
While accurate for general motions, these generated conditional control images might degrade in some challenging expressions. However, it is sufficient for
the control modules to learn the embedded motion clue and adapt and correlate to more precise expressions and poses when supervised with ground-truth motions. 

\section{User Study}
\label{sec:user_study}
The quantitative evaluation presented in the main paper (Table 1) highlights the effectiveness of~\papername. Specifically, in the "Expression/Pose" category, both Face Vid2vid Plus and~\papername~significantly outperform other methods. To further elucidate the distinctions between Face Vid2vid Plus and~\papername, we conducted a user study to substantiate our claim that~\papername~enhances the expressiveness of the base reenactment method. We invited 31 participants from various regions worldwide, each completing five questions following the format depicted in Figure~\ref{fig:user_study}. Participants were instructed to select synthesized videos that closely resembled the expression in the driving video. From the collected 155 responses, 129 out of 155 ($83.23\%$) affirmed that~\papername~surpasses Face Vid2vid Plus in terms of expressions and poses.

}
\bibliography{main}